\documentclass[10pt,a4paper]{article}
\usepackage{osajnl}

\usepackage{amsmath}
\usepackage{amssymb}
\usepackage{amsfonts}
\usepackage{graphicx}

\DeclareMathAlphabet{\mathitbf}{OT1}{cmr}{bx}{it} % italic bold
\newcommand{\ivec}[1]{\mathitbf{#1}}
\newcommand{\imat}[1]{\mathitbf{#1}}
\newcommand{\oned}{\mbox{\textsc{1-d}}\ }
\newcommand{\twod}{\mbox{\textsc{2-d}}\ }
\newcommand{\threed}{\mbox{\textsc{3-d}}\ }
\newcommand{\tp}{{\scriptscriptstyle\top}} % transpose

\newcommand{\lf}{\ell} % left
\newcommand{\rt}{r} % right

\begin{document}

\title{Cyclopean Geometry of Binocular Vision}

\author{Miles Hansard and Radu Horaud}

\affiliation{INRIA Rh\^{o}ne-Alpes, 655 Avenue de l'Europe,\\ 38330 Montbonnot, France.}

\begin{abstract}
The geometry of binocular projection is analyzed, with reference to
the primate visual system. In particular, the effects of coordinated
eye movements on the retinal images are investigated. An appropriate 
oculomotor parameterization is defined, and is shown to complement
the classical version and vergence angles. The midline horopter is 
identified, and subsequently used to construct the epipolar geometry 
of the system. It is shown that the Essential matrix can be obtained 
by combining the epipoles with the projection of the midline horopter.
A local model of the scene is adopted, in which depth is measured 
relative to a plane containing the fixation point. The binocular 
disparity field is given a symmetric parameterization, in which the 
unknown scene-depths determine the location of corresponding 
image-features. The resulting Cyclopean depth-map can be combined 
with the estimated oculomotor parameters, to produce a local 
representation of the scene. The recovery of visual direction and 
depth from retinal images is discussed, with reference to the 
relevant psychophysical and neurophysiological literature.
\end{abstract}


\section{Introduction}
\label{sec:introduction}

Information about the \threed structure of a scene can be recovered from binocular 
image-pairs, owing to the spatial separation of the two viewpoints. 
If the projections of corresponding points can be identified in the left and right 
images, then in principle, the scene can be reconstructed from the resulting 
binocular disparity field. In addition to the image information, the reconstruction
process requires knowledge (or estimation) of the projection parameters; in particular, 
the relative orientation of the eyes must be known. If some, or all, of these calibration
parameters remain unknown, then reconstruction process is limited to affine or projective 
properties of the scene\cite{faugeras-1995,luong-1996}. Hence, in order to facilitate the recovery of 
as much information as possible from the disparity field, it would be desirable to keep the 
orientation of the eyes fixed in the head. Despite this fact, human vision involves 
frequent eye movements, which serve a variety of purposes\cite{carpenter-1988}. 
For example, the eyes may be moved in order to direct the field of view, or to 
foveate an object of interest. Eye movements are also used to stabilize the retinal 
image with respect to head movements\cite{walls-1961}, and to track moving visual 
targets. It would be undesirable to suspend these functions, which are essentially
monocular, during the binocular analysis of a scene. 

The geometry of binocular stereopsis is complicated by movements of the eyes, as
described above. However, the two eyes typically move in a coordinated fashion, such 
that a single point in the scene is fixated. This can be achieved, in particular, 
by vergence eye-movements, which are \emph{driven} by binocular disparity.\cite{rashbass-1961} 
These coordinated eye movements benefit steropsis, as they
align the two retinal images at the respective foveas. It follows that the amount
of disparity around the fixation point tends to be reduced, assuming that the scene is
locally smooth. This is important, given the relatively short range of biological
disparity detectors.\cite{fleet-1996,prince-2002}

There may also be ethological reasons for the existence of binocular eye-movements, 
despite the resulting complication of stereopsis. It has been suggested that the evolution 
of binocular vision was motivated by the ability to detect camouflaged prey, by segmentation 
in depth, with respect to the background \cite{julesz-1971}. Another impetus may have
been the improvement in image quality that can be achieved by combining two views, especially
in nocturnal conditions \cite{pettigrew-1986}. Both of these applications would
benefit from binocular eye movements, which allow the scene to be scanned without moving the 
head, and help to register the two images. Because neither application requires geometric 
reconstruction of the scene, the disadvantages of moving the eyes are negligible. It 
should, however, be noted that stereopsis also exists in animals that do not move their 
eyes significantly, such as owls.\cite{pettigrew-1979}

It is clear that the binocular vision of humans (and other primates) has evolved beyond
simple tasks such as camouflage-breaking. Psychophysical evidence shows that at least 
some geometric properties of a \threed layout can be estimated by stereopsis, and that
these estimates can be combined, as the eyes fixate successive points in the visible
scene\cite{foley-1980}. Furthermore, it is clear that most types of eye movement are 
binocularly coordinated\cite{carpenter-1988}. 

The combination of eye movements and stereopsis raises important questions about
the oculomotor parameterization, the processing of the binocular disparity field, and the 
representation of the scene. These three questions are developed in more
detail below, in sections~\ref{sec:introduction-a}, \ref{sec:introduction-b}
and \ref{sec:introduction-c}, respectively.

\subsection{Oculomotor Parameterization}
\label{sec:introduction-a}

Firstly, what is the appropriate parameterization of binocular eye movements? This 
an important issue, because it determines the complexity of the control problem 
that the oculomotor system must solve. In particular, it is desirable to establish
the minimal number of parameters that are compatible with the observed range of 
oculomotor behaviour. The combination of two sensors, each of which can rotate in space, 
results in a system which has six angular degrees of freedom. 
However, if Donder's Law is obeyed\cite{carpenter-1988}, then the rotation of each 
eye \emph{around} the corresponding line of sight is determined by the \emph{direction} of that
line of sight. This removes one degree of freedom from each eye. Furthermore, binocular fixation 
implies co-planarity of the visual axes, which removes one elevation angle from the 
parameterization. This leaves three degrees of freedom, which can be conveniently
assigned to the elevation, azimuth and distance of the fixation point. These variables
are most naturally specified in relation to the `Cyclopean point' \cite{helmholtz-1910}, which,
in the present work, is defined as the midpoint of the inter-ocular axis. 
The trigonometry of this Cyclopean parameterization is defined in section 
\ref{sec:binocular-orientation}, and its relationship to the classical 
version/vergence model\cite{hering-1868} is stated.

\subsection{Disparity Processing}
\label{sec:introduction-b}

The second question that is addressed here concerns the relationship between the orientation
of the eyes, and the structure of the binocular disparity field. The difference of position 
between the left and right projections of a given scene point is, by convention, called the 
`absolute disparity' of the point\cite{blakemore-1970}. This is the quantity that can be 
measured most directly, by disparity-sensitive mechanisms\cite{cumming-1999}. It is important
to note that the response of such a mechanism must depend on the orientation of the eyes. Indeed,
for a typical scene, and a typical fixation point, it may be hypothesized that the relative 
orientation of the eyes will be the dominant source of absolute disparity.

It is important, for the reasons given above, to establish exactly how the disparity field is 
affected by eye movements. The question is approached in section~\ref{sec:fixed-points}, in which
the horopter of the fixating system is defined; this is the set of scene points that project to 
the same location in each image.\cite{helmholtz-1910} The horopter is used in 
section~\ref{sec:epipolar-geometry} to construct the epipolar geometry\cite{longuet-higgins-1981} 
of the system, which is effectively parametrized by the version and vergence angles. If a projected 
point is identified in one image, then the epipolar constraint restricts the location of the 
corresponding point to a \emph{line} in the other image. This important relationship can be 
expressed for any configuration of the eyes. In principle, the epipolar geometry could be used
to `rectify' the retinal images, thereby removing the effect of eye movements on the disparity 
field. However, this would not be consistent with the observed dependence of early binocular 
processing on absolute retinal disparity\cite{cumming-1999}. Hence it is desirable to 
parameterize the disparity field, with respect to the orientation of the eyes. The epipolar
geometry is the basis of such a parameterization.

\subsection{Scene Representation}
\label{sec:introduction-c}

The two questions described above are part of a third, more general question: How can the
geometric structure of the scene be represented by the visual system? This issue is 
complicated by the fact that the early mechanisms of primate binocular vision are sensitive to 
a quite limited range of disparities\cite{fleet-1996,prince-2002}. The region of space that can be 
resolved in depth depends, consequently, on the relative orientation of the eyes. 
Specifically, only those points in Panum's area (which is centred on the fixation point) 
can be fused\cite{ogle-1950,blakemore-1970}. It follows that any global representation of the  
scene must be assembled piecewise, over a series of fixations. It is natural to formulate this
process as the measurement of scene-structure with respect to a reference surface, followed
by an integration of the resulting local models\cite{bergen-1992}. The
plane that passes through the fixation point, and that is orthogonal to the Cyclopean visual
direction, is a convenient local model for binocular scene representation, as will be shown 
in section~\ref{sec:binocular-disparity}.

\subsection{Scope and Assumptions}
\label{sec:introduction-d}

The word `Cyclopean' has, in the present context, several possible meanings. As described in
section~\ref{sec:introduction-a}, the `Cyclopean point' is a notional centre of projection, located
at the midpoint of the inter-ocular axis (cf.\ Helmholtz\cite{helmholtz-1910}). It is convenient to 
use this point as the origin of the binocular coordinate system, although there is a useful alternative, 
as will be shown in section~\ref{sec:binocular-orientation}. The word `Cyclopean' is used elsewhere in a 
more general sense, with reference to visual information that is intrinsically binocular, such 
as the `edges' that can be perceived in a random-dot stereogram (cf.\ Julesz\cite{julesz-1971}). The 
phrase `Cyclopean geometry', as used here, refers to the fact that the binocular configuration of a 
fixating visual system can be parameterized by the state of a \emph{single} eye 
(cf.\ Hering\cite{hering-1868}), as will be explained in section~\ref{sec:binocular-orientation}.

It will be assumed, in this paper, that the retinal projections can be described by the usual 
pin-hole camera equations, and that these projections are `internally calibrated'. This means that the
visual system is able to relate the monocular retinal separation of any two points to the angle between 
the corresponding optical rays. It is unlikely that the visual system actually achieves such a Euclidean
representation \cite{todd-2001}. However, the present work is primarily a \emph{description} of binocular
geometry, which need not necessarily correspond to the encoding that is used by the visual system. The 
main constructions of this paper, including the horopter and the epipolar geometry, can be obtained 
(more naturally) in the projective setting, which does not require retinal calibration \cite{faugeras-1995,luong-1996}.
However, the present work is much concerned with the effects of the oculomotor configuration on binocular 
vision, and these effects are more readily studied by Euclidean methods. 

A distinction should be made between \emph{descriptions} and \emph{models} of binocular vision.
The present work aims to describe binocular geometry in the most convenient way. This 
leads to Cyclopean parameterizations of visual direction and binocular disparity. Whether these 
parameterizations are actually used by the visual system is a further question\cite{julesz-1972}.
In particular, it is not \emph{assumed} here that the Cyclopean representation has any biological 
reality. Discussion of the psychophysical and physiological evidence that can be used to make such
claims is confined to section~\ref{sec:discussion}.

It will be assumed here that the left and right visual axes intersect at a point in space. This is 
approximately true in practice. In the absence of an intersection, it would be possible to define an 
appropriate chord between the left and right visual axes, and to choose a notional fixation point on 
this segment. Finally, it is assumed that each eye rotates in accordance with Donders' law, meaning
that the cyclotorsion of the eyes can be estimated from the gaze direction.\cite{carpenter-1988}

\subsection{Novel Contributions}
\label{sec:introduction-e}

The results that are presented here are related to established approaches in computer vision
\cite{faugeras-1995,luong-1996,armstrong-1995,shashua-navab-1996}. The derivations however, 
are novel, and the details are specific to the biological context. The following results are 
of particular interest: \
I.~The Cyclopean parameterization of
binocular orientation (equations \ref{eqn:p0},\ref{eqn:v-ab},\ref{eqn:beta}); \
II.~The identification of the midline horopter as an axis that passes through the pole of 
the visual plane (\ref{eqn:ca},\ref{eqn:qa}); \
III.~The construction of the Essential matrix from the epipoles and midline horopter (\ref{eqn:e}); \
IV.~The symmetric parameterization of binocular correspondence (\ref{eqn:cyc-disparity}) and 
V.~the parameterization of binocular parallax as a function of deviation from the fixation plane
(\ref{eqn:parallax-3b}).

\section{Projection Model}
\label{sec:projection-model}

The notation and coordinate systems used in this paper are described here. Points and vectors will be 
shown in bold type, for example $\ivec{q}$, $\ivec{v}$. The transpose of $\ivec{v}$ is a row-vector 
$\ivec{v}^\tp$, and the Euclidean length is $|\ivec{v}|$. The notation $(\ivec{v})_3$ will be used to 
indicate the third component of the vector $\ivec{v}$. Matrices are represented by upper-case letters,
for example $\imat{M}$.

The \threed Euclidean coordinates of a point will be distinguished by a bar, e.g.\ $\bar{\ivec{q}}$. 
Note that the difference of two Euclidean points results in a vector, 
e.g.\ $\ivec{v} = \bar{\ivec{q}}-\bar{\ivec{p}}$. The homogeneous image-coordinates of points and lines
are written without a bar; for example, a point at retinal location
$(x,y)^\tp$ is represented by
$\ivec{q} = (\mu x,\mu y, \mu)^\tp$, with $\mu \ne 0$. Note that
the inhomogeneous coordinates can be recovered from $\ivec{q}/\mu$. Scalar multiples 
of the homogeneous coordinates represent the same image point. For example, if 
$\ivec{p} = (\lambda x,\lambda y,\lambda)^\tp$ then $\ivec{q}/\mu = \ivec{p}/\lambda$; this relationship 
will be written as $\ivec{p} \sim \ivec{q}$. 

A line in the image plane has homogeneous coordinates 
$\ivec{n} = (a,b,c)^\tp$, such that $\ivec{q}$ is on $\ivec{n}$ if $\ivec{n}^\tp\ivec{q} = 0$. Scalar 
multiples represent the same line; if $\ivec{m} = (\kappa a, \kappa b, \kappa c)^\tp$, then
 $\ivec{m} \sim \ivec{n}$, with $\ivec{m}^\tp\ivec{q} = 0$, as before. If $\ivec{n}$ is defined as 
$\ivec{n} = \ivec{p} \times \ivec{q}$, then $\ivec{n}^\tp\ivec{p} = \ivec{n}^\tp\ivec{q} = 0$; hence 
$\ivec{n}$ is the line through the two points. Similarly, if $\ivec{q} = \ivec{m}\times\ivec{n}$ 
then $\ivec{m}^\tp\ivec{q} = \ivec{n}^\tp\ivec{q}=0$; hence $\ivec{q}$ is the intersection point of the 
two lines. 

The left and right optical centres are labeled $\bar{\ivec{c}}_\lf$ and $\bar{\ivec{c}}_\rt$, respectively. 
The Cyclopean point $\bar{\ivec{c}}_b$ is fixed halfway between, and it will also be useful to define the 
`baseline' vector $\ivec{b}$;
\begin{align}
\ivec{b} &= \bar{\ivec{c}}_\rt-\bar{\ivec{c}}_\lf \label{eqn:b}\\
\bar{\ivec{c}}_b &= \tfrac{1}{2}(\bar{\ivec{c}}_\lf+\bar{\ivec{c}}_\rt) \label{eqn:c}.
\end{align}
Only the \emph{ratio} of the scene size to the baseline length can be recovered from the images, in the 
absence of other information. For this reason it is helpful to define the distance between the two optical 
centres as $|\ivec{b}|=1$, so that Euclidean coordinates are measured units of interocular separation. 
The location of the scene coordinate-system is immaterial, so it will be convenient to put the origin 
at the Cyclopean point $\bar{\ivec{c}}_b$. The coordinates of the optical centres, with reference to 
figure~\ref{fig:visplane}, will be
\begin{equation}
\bar{\ivec{c}}_\lf = -\bigl(\tfrac{1}{2}, 0, 0\bigr)^\tp
\quad\text{and}\quad
\bar{\ivec{c}}_\rt = \bigl(\tfrac{1}{2}, 0, 0\bigr)^\tp.
\label{eqn:cl,cr}
\end{equation}
The baseline vector (\ref{eqn:b}) is therefore parallel to the $\ivec{x}$ axis, and a perpendicular axis 
$\ivec{z} = (0,0,1)^\tp$ will be taken as the head-centric outward direction. These two vectors define 
Cartesian coordinates in the horizontal plane. The downward normal of this plane is $\ivec{y} = (0,1,0)^\tp$, 
so that the axes $\ivec{x}$, $\ivec{y}$ and $\ivec{z}$ form a right-handed system, as shown in 
figure~\ref{fig:visplane}. 

The orientations of the left, right and Cylcopean eyes are expressed by $3\times 3$ rotation matrices 
$\imat{R}_\lf$, $\imat{R}_\rt$ and $\imat{R}$, respectively. A view of the scene is obtained by expressing 
each point $\bar{\ivec{q}}$ relative to an optical centre, and applying the corresponding rotation. 
Hence the homogeneous perspective projection into the left image is
\begin{equation}
\ivec{p}_\lf \sim \imat{R}_\lf (\bar{\ivec{q}} - \bar{\ivec{c}}_\lf).
\label{eqn:projection}
\end{equation}
If both eyes are looking straight ahead (with zero cyclo-rotation),
$\imat{R}_\lf = \imat{R}_\rt = \imat{I}$, then the 
projections of $\bar{\ivec{q}}=(x,y,z)^\tp$ can be computed easily; they are
$
\ivec{q}_\lf \sim \bigl(x+\tfrac{1}{2}, y, z\bigr)^\tp
$
and
$
\ivec{q}_\rt \sim \bigl(x-\tfrac{1}{2}, y, z\bigr)^\tp
$.
The inhomogeneous coordinates of these points are $x_\lf = (x+\tfrac{1}{2}) \big/ z$, 
$x_\rt = (x-\tfrac{1}{2}) \big/ z$, and $y_\lf = y_\rt = y / z$, corresponding 
to the ordinary perspective equations. It is straightforward, in this case, to evaluate the 
\emph{binocular disparity} $(x_\lf,y_\lf)^\tp - (x_\rt,y_\rt)^\tp = (1/z,0)^\tp$. Note that,
in this case of parallel visual axes, the disparity vector is confined to the horizontal direction.
For general orientations of the eyes, disparity equations are more complicated, as will be seen in 
section~\ref{sec:binocular-disparity}.

\section{Binocular Orientation}
\label{sec:binocular-orientation}

An angular parameterization of binocular eye movements is introduced in this section.
As was noted in section \ref{sec:introduction-a}, the degrees of freedom of the 
binocular system can be reduced from six, to three. The reduction is achieved by imposing
the fixation constraint, together with Donders' Law. An appropriate parameterization will
be developed, based on the Cyclopean azimuth, elevation and distance of the fixation point.
It will be shown that this representation complements the classical version/vergence 
coordinates\cite{hering-1868}.

\begin{figure}[!ht]
\begin{center}
\includegraphics[scale=1,angle=0]{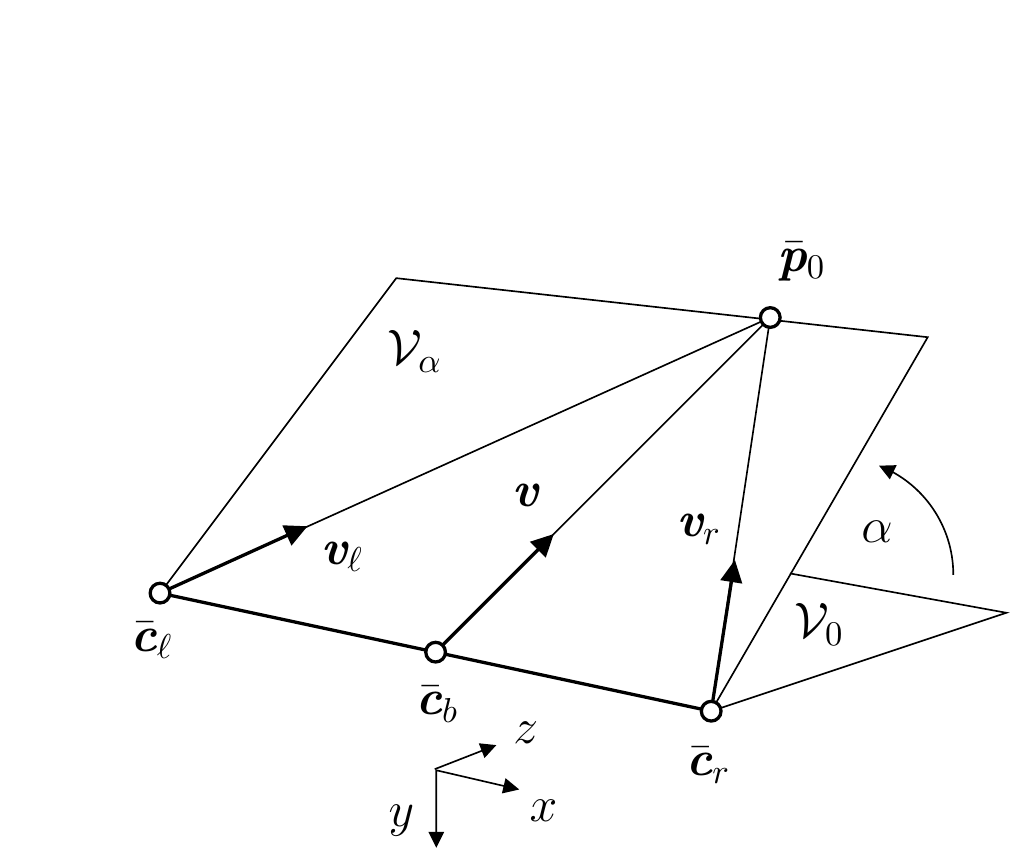}
\caption{A visual plane $\mathcal{V}_\alpha$ is defined by the optical centres $\bar{\ivec{c}}_\lf$ and $\bar{\ivec{c}}_\rt$, together with the fixation point $\bar{\ivec{p}}_0$. The visual
directions $\ivec{v}$, $\ivec{v}_\lf$ and $\ivec{v}_\rt$ lie in this plane, which has 
an elevation angle $\alpha$.
The scene coordinates are located at the Cyclopean point $\bar{\ivec{c}}_b = (0,0,0)^\tp$, such that $\mathcal{V}_0$ coincides with the $x,z$ plane.}
\label{fig:visplane}
\end{center}
\end{figure}

Suppose that the point $\bar{\ivec{p}}_0$ is to be fixated. The scene coordinates of this
point can be specified by a head-fixed direction $\ivec{v}$ from the Cyclopean origin, 
in conjunction with a distance, $\rho$, along the corresponding ray;
\begin{equation}
\bar{\ivec{p}}_0 = \rho \ivec{v}.
\label{eqn:p0}
\end{equation}
The direction $\ivec{v}$ is a unit-vector, and the positive scalar $\rho$ will be called the 
\emph{range} of the fixation point $\bar{\ivec{p}}_0$. The Cyclopean direction may be written in 
terms of the elevation and azimuth angles $\alpha$ and $\beta$ respectively;
\begin{equation}
\ivec{v} = (\sin\beta,\, -\sin\alpha\, \cos\beta,\, \cos\alpha\, \cos\beta)^\tp
\label{eqn:v-ab}
\end{equation}
where $\cos\beta$ is the projected length of $\ivec{v}$ in the mid-sagittal plane $x=0$, which divides 
one side of the head from the other. Note that the 
elevation $\alpha$ is positive for points above the horizontal plane $(y < 0)$, and that the azimuth 
$\beta$ is positive for points to the right of the mid-sagittal plane $(x > 0)$. These visual angles will 
each be in the range $[-\pi/2,\pi/2]$, so that any point with $z \ge 0$ can be identified, as shown 
in figure \ref{fig:visplane}. If the fixation point $\bar{\ivec{p}}_0 = (x,y,z)^\tp$ is given in 
Cartesian coordinates, then the corresponding range and direction are 
\begin{align}
\rho &= |\bar{\ivec{p}}_0| \label{eqn:rho}\\
\ivec{v} &= \bar{\ivec{p}}_0/\rho \label{eqn:v}
\end{align}
respectively. The elevation and Cyclopean azimuth angles can be obtained from the equations 
$\tan\alpha = -y / z$ and $\sin\beta = x/\rho$ respectively. The vector $(\alpha,\beta,\rho)^\tp$ 
contains the \emph{Helmholtz coordinates} of the point $\bar{\ivec{p}}_0$.

In addition to the Cyclopean visual axis $\ivec{v}$, defined above (\ref{eqn:v}), there exist left and right
axes $\ivec{v}_\lf$ and $\ivec{v}_\rt$, respectively. If the eyes are fixating the point $\bar{\ivec{p}}_0$, as described above, then $\ivec{v}_\lf$ and $\ivec{v}_\rt$ can be derived from $\ivec{v}$ and $\rho$, as will be shown below.
The optical centres $\bar{\ivec{c}}_\lf$ and $\bar{\ivec{c}}_\rt$,
together with the fixation point $\bar{\ivec{p}}_0$ define a \emph{visual plane}, $\mathcal{V}_\alpha$ as shown in fig.\ \ref{fig:visplane}.
The three visual axes intersect at $\bar{\ivec{p}}_0$, and so $\ivec{v}_\lf$, $\ivec{v}_\rt$ and $\ivec{v}$ lie in $\mathcal{V}_\alpha$.
All of the possible visual planes contain the baseline $\ivec{b}$, and may be parameterized by the dihedral
angle $\alpha$ between $\mathcal{V_\alpha}$ and the horizontal plane $\mathcal{V}_0$. The azimuth angles $\beta$, $\beta_\lf$ and $\beta_\rt$ will now be defined \emph{in} the visual plane $\mathcal{V}_\alpha$.

\begin{figure}[!ht]
\begin{center}
\includegraphics[scale=1,angle=0]{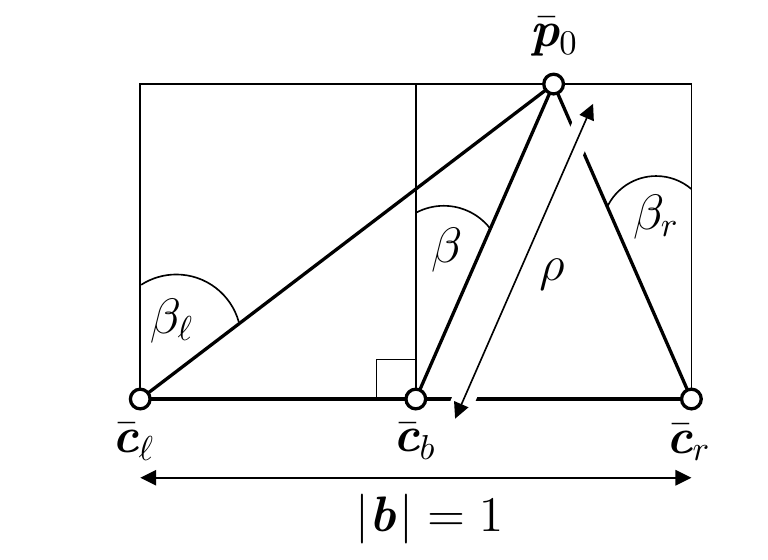}
\caption{Angular coordinates of a point $\bar{\ivec{p}}_0$ in the visual plane $\mathcal{V}_\alpha$. The optical centres are indicated by $\bar{\ivec{c}}_\lf$ and $\bar{\ivec{c}}_\rt$. The azimuth angles 
$\beta$ and $\beta_\lf$ are positive in this example, whereas $\beta_\rt$ is negative. The
Cyclopean range of the fixation point is $\rho$.}
\label{fig:angular}
\end{center}
\end{figure}

Firstly it will be shown that, if the eyes are fixating, then the left and right visual
directions can be simultaneously parameterized by the Cyclopean direction and distance
of the fixation point.
It is convenient to begin by assuming that the fixation point is in the horizontal plane,
such that $\alpha=0$. The removal of this assumption will be discussed subsequently.
It can be seen, with reference to figure \ref{fig:angular}, that if the baseline separation is 
$|\ivec{b}|=1$, then $\tan\beta_\lf$ and $\tan\beta_\rt$ are equal to 
$(\rho\sin\beta \pm \tfrac{1}{2}) / (\rho\cos\beta)$. Some re-arrangement leads 
to the definitions
\begin{equation}
\tan \beta_\lf = \tan\beta + \frac{\sec\beta}{2\rho}
\quad\text{and}\quad 
\tan \beta_\rt = \tan\beta - \frac{\sec\beta}{2\rho}.
\label{eqn:beta}
\end{equation}
It is clear from these equations that, for a given Cyclopean azimuth $\beta$, the visual 
directions become more equal as the fixation distance, $\rho$, increases. It may also be 
noted that if $\beta=0$, then the fixation is symmetric, with left and right azimuths 
$\pm \tan^{-1} (\tfrac{1}{2} / \rho)$, as is commonly assumed in the literature.

If the fixation point $\bar{\ivec{p}}_0$ is in $\mathcal{V}_\alpha$, then the matrices 
representing the orientation of the eyes are easily constructed. For example, the matrix 
$\imat{R}_\lf$ in (\ref{eqn:projection}) is
\begin{equation}
\imat{R}_\lf = 
\begin{pmatrix}
\cos\beta_\lf & 0 & -\sin\beta_\lf\\
0 & 1 & 0\\
\sin\beta_\lf & 0 & \cos\beta_\lf
\end{pmatrix}.
\label{eqn:r-l}
\end{equation}
The analogous definitions are made for the matrices $\imat{R}$ and $\imat{R}_\rt$, with 
angles $\beta$ and $\beta_\rt$, respectively.

Although the Helmholtz coordinates are convenient for specifying visual directions, the eyes do not, in
general, rotate around the corresponding axes. One characteristic of actual eye movements is that, for 
general fixation points, each eye will be cyclo-rotated around the corresponding visual 
direction. Although the cyclo-rotation angles $\gamma_\lf$ and $\gamma_\rt$ are non-zero, Donders' law
states that they are completely determined by the corresponding visual directions; hence there
exist functions $\gamma_\lf(\alpha,\beta_\lf)$ and $\gamma_\rt(\alpha,\beta_\rt)$. The definition
of these functions can be obtained from, for example, Listing's law and its extensions.\cite{hepp-1994}

The effects of cyclo-rotation are relevant to the distribution of physiological 
disparity mechanisms across the visual field \cite{koenderink-1976}. However, cyclo-rotation 
is of limited interest with respect to the analysis of binocular \emph{geometry};
this is because the parameters $\gamma_\lf$ and $\gamma_\rt$ simply determine
rotations of the respective \emph{images}, independently of any \threed information.
For this reason, the effects of cyclo-rotation will not be incorporated into
the following derivations, meaning that Donders' law is obeyed in the trivial sense;
\begin{equation}
\gamma_\lf(\alpha,\beta_\lf) = \gamma_\rt(\alpha,\beta_\rt) = 0.
\label{eqn:gamma}
\end{equation}
It is, at the end of all derivations, easy to substitute any other example of Donders'
law, as will be explained in sections \ref{sec:fixed-points} and \ref{sec:binocular-disparity}. The practical advantage of 
the restriction (\ref{eqn:gamma}) is that any dependence on the elevation angle  $\alpha$
is removed from the analysis. This makes it possible to study the binocular geometry with 
respect to fixation points in a \emph{single} visual plane, without loss of generality. 
Furthermore, Listing's law agrees with (\ref{eqn:gamma}) when $\alpha=0$; hence it is 
natural to choose the horizontal plane $\mathcal{V}_0$ for further investigation.

The remainder of this section will be concerned with the classical binocular \emph{vergence} 
and \emph{version} parameters, and their relation to the Helmholtz coordinates. This discussion 
will lead to the definition of binocularly \emph{fixed} points. Section \ref{sec:fixed-points} will develop 
this idea, which will subsequently be used to define the epipolar geometry.

\begin{figure}[!ht]
\begin{center}
\includegraphics[scale=1,angle=0]{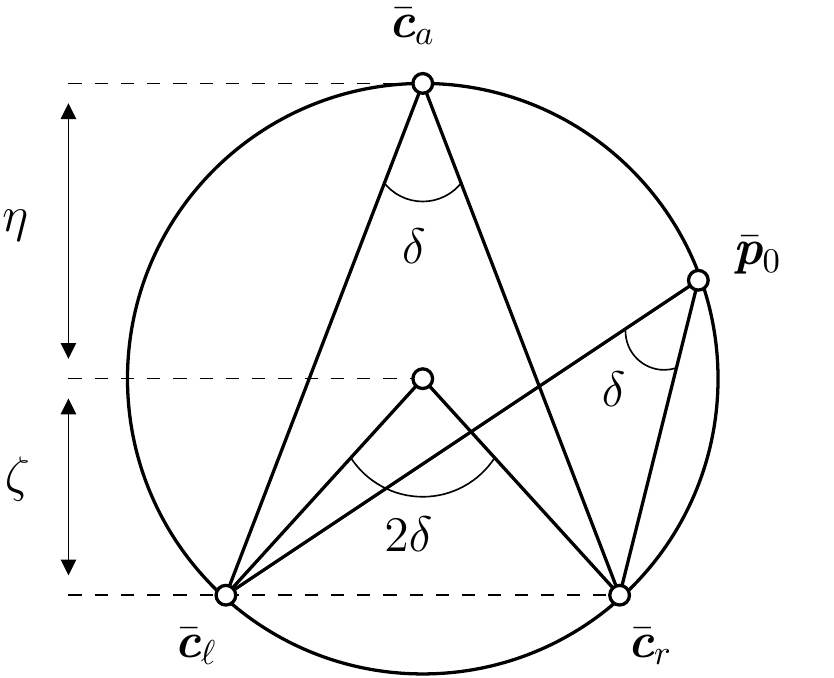}
\caption{The Vieth-M\"{u}ller circle is defined by the positions of the optical
centres $\bar{\ivec{c}}_\lf$ and $\bar{\ivec{c}}_\rt$, together with the fixation
point, $\bar{\ivec{p}}_0$.
The forward $(z>0)$ arc of the circle intersects the mid-sagittal plane at
the point $\bar{\ivec{c}}_a$.
The vergence angle, $\delta$, is inscribed at $\bar{\ivec{p}}_0$
by $\bar{\ivec{c}}_\lf$ and $\bar{\ivec{c}}_\rt$. The same angle is inscribed at all 
other points on the circle, including $\bar{\ivec{c}}_a$.}
\label{fig:vergence}
\end{center}
\end{figure}

The vergence angle, $\delta$, will be defined as the angle between the lines of sight at the 
fixation point; the version angle, $\epsilon$, will be defined as the average gaze azimuth.
In relation to the Helmholtz coordinates, this means that
\begin{align}
\delta &= \beta_\lf - \beta_\rt \label{eqn:delta}\\
\epsilon &= \tfrac{1}{2}(\beta_\lf + \beta_\rt). \label{eqn:epsilon}
\end{align}
The vergence angle $\delta$ is non-negative, due to the inequality 
$\beta_\rt \le \beta_\lf$, which follows from the signs and limits of $\beta_\lf$ and 
$\beta_\rt$ as defined above. The equality $\beta_\lf=\beta_\rt$ occurs for infinitely 
distant fixation points, for which $\delta=0$. These definitions are illustrated in 
figure \ref{fig:vergence}.

%The version angle $\epsilon$ is comparable to the Cyclopean gaze azimuth $\beta$, but the 
%two are not, in general, equal. 

The properties of the vergence and version parameters can be understood with reference
to the Vieth-M\"{u}ller circle\cite{tyler-1991}, which is defined by the
two optical centres $\bar{\ivec{c}}_\lf$ and $\bar{\ivec{c}}_\rt$, together with the fixation
point $\bar{\ivec{p}}_0$. The vergence, $\delta$, is the inscribed angle at $\bar{\ivec{p}}_0$,
being opposite the inter-ocular axis $\ivec{b}$. The law of sines gives the diameter of the 
circumcirle as $1/\sin\delta$, with $|\ivec{b}|=1$, as usual. The angle subtended by $\ivec{b}$
from the centre of the circle is $2\delta$, being twice the inscribed angle. The isosceles 
triangle formed by $\bar{\ivec{c}}_\lf$, $\bar{\ivec{c}}_\rt$ and the centre of the circle can be split 
into two right-angled triangles, such that $\tan\delta= \frac{1}{2} / \zeta$, where $\zeta$ is
the $z$ coordinate of the centre. It follows that the Vieth-M\"{u}ller circle is centred at 
the point $(0,0,\zeta)^\tp$, and has radius $\eta$, where
\begin{align}
\zeta &= \tfrac{1}{2} \cot \delta \label{eqn:zeta}\\
\eta &= \tfrac{1}{2} \csc \delta. \label{eqn:eta}
\end{align}
The optical centres, $\bar{\ivec{c}}_\lf$ and $\bar{\ivec{c}}_\rt$, divide the Vieth-M\"{u}ller
circle into two arcs, according to the sign of $z$. The forward ($z\ge 0$) arc contains the 
fixation point, $\bar{\ivec{p}}_0$, with inscribed angle $\delta$. Furthermore, the inscribed 
angles at all other points $\bar{\ivec{q}}_\mathrm{\scriptscriptstyle VM}$ on this arc must
be equal; hence the Vieth-M\"{u}ller circle contains the locus of iso-vergence.

\begin{figure}[!ht]
\begin{center}
\includegraphics[scale=1,angle=0]{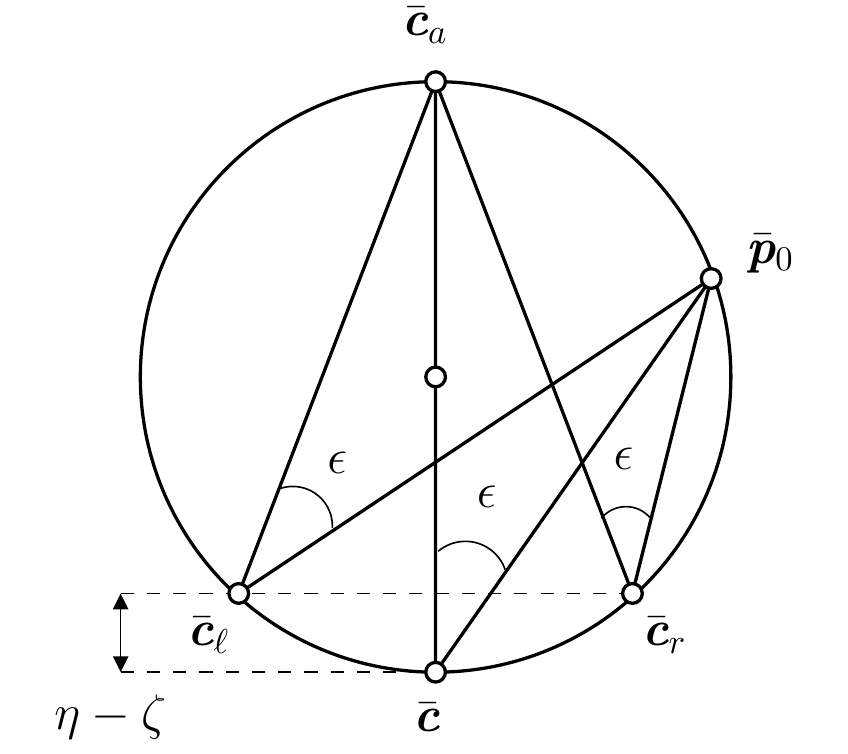}
\caption{The points $\bar{\ivec{c}}_a$ and $\bar{\ivec{p}}_0$ inscribe the version
angle, $\epsilon$, at an optical centre, $\bar{\ivec{c}}$, which is located on the 
backward $(z<0)$ arc of the Vieth-M\"{u}ller circle.
The same angle is inscribed at $\bar{\ivec{c}}_\lf$ and $\bar{\ivec{c}}_\rt$. It 
follows that as $\bar{\ivec{p}}_0$ is fixated, $\bar{\ivec{c}}_a$ lies in the same
visual \emph{direction} from each eye. Furthermore, the triangle defined by 
$\bar{\ivec{c}}_\lf$, $\bar{\ivec{c}}_\rt$, and 
$\bar{\ivec{c}}_a$ is isosceles, and so the point $\bar{\ivec{c}}_a$
is at the same \emph{distance} from each eye.}
\label{fig:version}
\end{center}
\end{figure}

The version angle $\epsilon$ gives the azimuth of $\bar{\ivec{p}}_0$ from a Cyclopean point 
$\bar{\ivec{c}} = (0,0,\zeta - \eta)^\tp$, which lies at the back of 
the Vieth-M\"{u}ller circle, as shown in figure \ref{fig:version}.
It may be noted that $\beta \approx \epsilon$, and that this approximation improves
as $\delta$ decreases. Furthermore, it can be shown that $\rho \approx 2/\eta\cos\epsilon$.
Evidently the location of the point $\bar{\ivec{c}}$ varies according to the 
vergence angle. This is one reason for \emph{deriving} the $(\delta,\epsilon)$ parameterization
from the $(\beta,\rho)$ parameterization, as above: The present analysis has a fixed reference 
point $\bar{\ivec{c}}_b=(0,0,0)^\tp$, meaning that visual information can easily be combined as the 
eyes re-fixate. Furthermore, the range parameter $\rho$ is directly related to the observer's behaviour, including listening and moving, whereas $\delta$ is specific to the oculomotor 
system. Nonetheless, the vergence and version parameters are crucial, as will be shown 
in the following sections. 

\section{Fixed Points}
\label{sec:fixed-points}

In this section it will be shown that, for a given vergence angle, certain points in the 
scene project to the same location in each image. These points constitute the geometric 
\emph{horopter} of the fixating system. The constructions that are given here will be used
to construct the epipolar geometry of the two images, in section \ref{sec:epipolar-geometry}.
For this purpose, it will be convenient to study the horopter in the absence of any
cyclo-rotation.

It was shown in section \ref{sec:binocular-orientation} that the Vieth-M\"{u}ller circle is 
defined by the optical centres $\bar{\ivec{c}}_\lf$ and $\bar{\ivec{c}}_\rt$, together with 
the fixation point $\bar{\ivec{p}}_0$. Consider another scene-point point,
$\bar{\ivec{q}}_\mathrm{\scriptscriptstyle VM}$, that lies on the forward section of the 
Vieth-M\"{u}ller circle. This point is in the visual plane $\mathcal{V}_0$, and therefore 
satisfies the equation $y=0$, as well as the conditions
\begin{equation}
\begin{split}
\begin{gathered}
x^2 + (z-\zeta)^2 = \eta^2\\
z \ge 0.
\end{gathered}
\end{split}
\label{eqn:vm}
\end{equation}
The two points $\bar{\ivec{p}}_0$ and $\ivec{\bar{q}}_\mathrm{\scriptscriptstyle VM}$,
both of which are on the forward section of the Vieth-M\"{u}ller circle, must inscribe 
equal angles at the optical centres. The point $\bar{\ivec{p}}_0$ is being 
fixated, and therefore appears in the fovea of each image. Hence the projected
point $\ivec{q}_\mathrm{\scriptscriptstyle VM}$ is fixed in the images; 
it appears on the horizontal meridian of each retina, at a fixed angular 
offset from the corresponding fovea.

The preceding argument has established that all scene points $\bar{\ivec{q}}_\mathrm{\scriptscriptstyle VM}$ are fixed in the images. This means, in camera coordinates, that the equations
$x_\lf/z_\lf = x_\rt/z_\rt$ and $y_\lf/z_\lf = y_\rt/z_\rt$ are satisfied. The
existence of points that are fixed in the scene, as well as in the images, will
now be investigated. 
Recall that the Euclidean coordinates of $\bar{\ivec{q}} = (x,y,z)^\tp$ in the left and right camera
frames are
$\bar{\ivec{q}}_\lf = \imat{R}_\lf(\bar{\ivec{q}} - \bar{\ivec{c}}_\lf)$ and
$\bar{\ivec{q}}_\rt = \imat{R}_\rt(\bar{\ivec{q}} - \bar{\ivec{c}}_\rt)$, respectively. 
The point $\bar{\ivec{q}}$ is fixed in the scene if $\bar{\ivec{q}}_\lf = \bar{\ivec{q}}_\rt$,
which in turn implies that $|\bar{\ivec{q}}_\lf|^2 = |\bar{\ivec{q}}_\rt|^2$.
The squared-lengths are preserved by the rotation matrices $\imat{R}_\lf$ and $\imat{R}_\rt$, 
and so $|\bar{\ivec{q}} - \bar{\ivec{c}}_\lf|^2 = |\bar{\ivec{q}} - \bar{\ivec{c}}_\rt|^2$.
From the definition of $\bar{\ivec{c}}_\lf$ and $\bar{\ivec{c}}_\rt$ in (\ref{eqn:cl,cr}), this is 
equivalent, in scene-coordinates, to 
the condition $(x+\tfrac{1}{2})^2 = (x-\tfrac{1}{2})^2$. Hence it can be seen
that any fixed point in the scene must lie in the mid-sagittal plane $x=0$, that 
divides one side of the head from the other. Substituting $x=0$ into
(\ref{eqn:vm}) leads immediately to $z = \zeta + \eta$, leaving $y$ free to vary. 
In general, $y_\lf = y_\rt$, because the axis of the vergence rotation
is perpendicular to the visual plane $\mathcal{V}_\alpha$. This argument has
established that there is an axis of fixed points $\bar{\ivec{q}}_a$ in the
scene. This axis intersects the Vieth-M\"{u}ller circle at a point 
$\bar{\ivec{c}}_a$, and is perpendicular to the visual plane. If, without
loss of generality, $\alpha=0$, then the coordinates of these points are
\begin{align}
\bar{\ivec{c}}_a &= (0,0,\zeta+\eta)^\tp \label{eqn:ca}\\
\bar{\ivec{q}}_a &= \bar{\ivec{c}}_a + (0,y,0)^\tp. \label{eqn:qa} 
\end{align}
This axis of points, $\bar{\ivec{q}}_a$, which has been identified elsewhere\cite{tyler-1991,cooper-1979},
is the geometric \emph{midline horopter}. The point $\bar{\ivec{c}}_a$ is the \emph{pole}
of the planar transformation induced by the translation $\ivec{b}$ and vergence
rotation $\imat{R}_\rt\imat{R}_\lf^\tp$. The points $\bar{\ivec{q}}_a$ lie on 
the associated screw-axis\cite{armstrong-1995}.

It will be useful to compute the \emph{image} coordinates of the axis, which are common
to both eyes, as shown in fig.\ \ref{fig:version}. The points $\bar{\ivec{p}}_0$ and $\bar{\ivec{c}}_a$ inscribe equal angles
at $\bar{\ivec{c}}_\lf$, $\bar{\ivec{c}}_\rt$ and 
$\bar{\ivec{c}}$; moreover, the angle at 
$\bar{\ivec{c}}$ is, by definition, the binocular version,
$\epsilon$. Having established that the angular direction of $\bar{\ivec{q}}_a$ from 
either optical centre is $\epsilon$, the common distance of this point will also be 
computed. The points $\bar{\ivec{c}}_\lf$, $\bar{\ivec{c}}_\rt$ and 
$\bar{\ivec{c}}_a$ form an isosceles triangle, from which it 
can be seen that $|\bar{\ivec{c}}_a| \sin(\delta/2) = \frac{1}{2}$. 
It follows that, in the coordinates of either eye, the axis is specified by
\begin{align}
\ivec{c}_a &= \tfrac{1}{2}\csc(\delta/2) \, (-\sin\epsilon, 0, \cos\epsilon)^\tp \label{eqn:ca-img}\\
\ivec{q}_a &= \ivec{c}_a + (0,y,0)^\tp. \label{eqn:qa-img}
\end{align}
These image points lie on a vertical line $\ivec{a}$, which has the same coordinates in 
each eye. The equation of the line is $\ivec{q}_a^\tp\ivec{a} = 0$, and so it follows from 
equation (\ref{eqn:ca-img}) that the coordinates of the line are determined by the 
version angle $\epsilon$;
\begin{equation}
\ivec{a} \sim (\cos\epsilon, 0, -\sin\epsilon)^\tp. \label{eqn:a}
\end{equation}
The results of this section can be summarized as follows.
The geometric horopter, in the absence of cyclo-rotation, consists of the forward 
part of the Vieth-M\"{u}ller circle, together with the midline component.
If a scene point $\bar{\ivec{q}}$ is on the horopter, then the image
coordinates of the corresponding points are equal, $\ivec{q}_\lf \sim \ivec{q}_\rt$.
Furthermore, if the point
is on the midline component, then it is also true that the eye-coordinates of $\bar{\ivec{q}}$
are equal, $\bar{\ivec{q}}_\lf = \bar{\ivec{q}}_\rt$. Finally, the image coordinates
(\ref{eqn:a}) of the midline are determined by the binocular version angle.
It should be noted that, in the presence of cyclo-rotation, the geometric horopter
takes the form of a twisted cubic curve\cite{helmholtz-1910}. This curve coincides
with the Vieth-M\"{u}ller circle as it passes through the optical centres, and has
asymptotes at $\bar{\ivec{c}}_a \pm (0,y,0)^\tp$.

%The axis is perpendicular to $\mathcal{V}$, and meets the plane at $\bar{\ivec{q}}_a$,
%which is the foremost point on the Vieth-M\"{u}ller circle. 
%The same result can be obtained directly, by solving the equation 
%$
%\imat{R}_\lf(\bar{\ivec{q}} - \bar{\ivec{c}}_\lf) =
%\imat{R}_\rt(\bar{\ivec{q}} - \bar{\ivec{c}}_\rt)
%$
%for scene-coordinates $\bar{\ivec{q}}$. If $\imat{S}=\imat{R}_\rt^\tp\imat{R}_\lf$, 
%then it can be shown that the solution is 
%$\bar{\ivec{q}}_a = (\imat{I}-\imat{S})^\dagger(\imat{I} + \imat{S}) \ivec{c}_\rt$, 
%where `$\dagger$' indicates the pseudo-inverse, and $(0,1,0)^\tp$ is a basis
%for the null-space of the rank-two matrix $(\imat{I}-\imat{S})$.

\section{Epipolar Geometry}
\label{sec:epipolar-geometry}

It was established, in the preceding section, that certain scene points have the
same coordinates in both images. The related \emph{epipolar constraint} is weaker,
but much more useful, as it applies to \emph{all} scene points. The epipolar geometry
of the fixating system will now be described; in particular, the image of the midline horopter 
(\ref{eqn:a}), will be used to construct the appropriate `Essential matrix'.\cite{longuet-higgins-1981}

The epipolar constraint is as follows: Given an image point $\ivec{q}_\lf$ in 
$\mathcal{I}_\lf$, the corresponding point $\ivec{q}_\rt$ in $\mathcal{I}_\rt$ 
must lie on a known epipolar line $\ivec{u}_\rt$, such that
$
\ivec{u}_\rt^\tp \! \ivec{q}_\rt = 0.
$
The geometric interpretation of this is that the scene point $\bar{\ivec{q}}$ must be 
located on the ray defined by the optical centre $\bar{\ivec{c}}_\lf$ and the image point
$\ivec{q}_\lf$; the ray projects to a line $\ivec{u}_\rt$ in the other view, and so 
$\ivec{q}_\rt$, being another image of $\bar{\ivec{q}}$, must lie on the line. 
Furthermore, note that the optical centre $\bar{\ivec{c}}_\lf$ is common to all such
rays, and so the resulting lines $\ivec{u}_\rt$ must intersect at a single point in
$\mathcal{I}_\rt$. This point is the right \emph{epipole}, $\ivec{e}_\rt$.
Similar arguments can be used to introduce the left epipole $\ivec{e}_\lf$, as well
as the associated lines $\ivec{u}_\lf$ in $\mathcal{I}_\lf$.

%The epipolar lines will now be constructed, with the aid of the following notation. The 
%three points $\bar{\ivec{c}}_\lf$, $\bar{\ivec{c}}_\rt$ and $\bar{\ivec{q}}$ define an 
%epipolar plane $\mathcal{U}_q$. The epipolar lines $\ivec{u}_\lf$ and $\ivec{u}_\rt$ are 
%the intersections of $\mathcal{U}_q$ with the image planes $\mathcal{I}_\lf$ and
%$\mathcal{I}_\rt$, respectively. In addition, the optical centres  $\bar{\ivec{c}}_\lf$
%and $\bar{\ivec{c}}_\rt$ project to the \emph{epipoles} $\ivec{e}_\rt$ and $\ivec{e}_\lf$
%in $\mathcal{I}_\rt$ and $\mathcal{I}_\lf$ respectively. Note that any epipolar line
%$\ivec{u}_\lf$ must pass through $\ivec{e}_\lf$, and likewise $\ivec{u}_\rt$ must pass
%through $\ivec{e}_\rt$.

\begin{figure}[!ht]
\begin{center}
\includegraphics[scale=1]{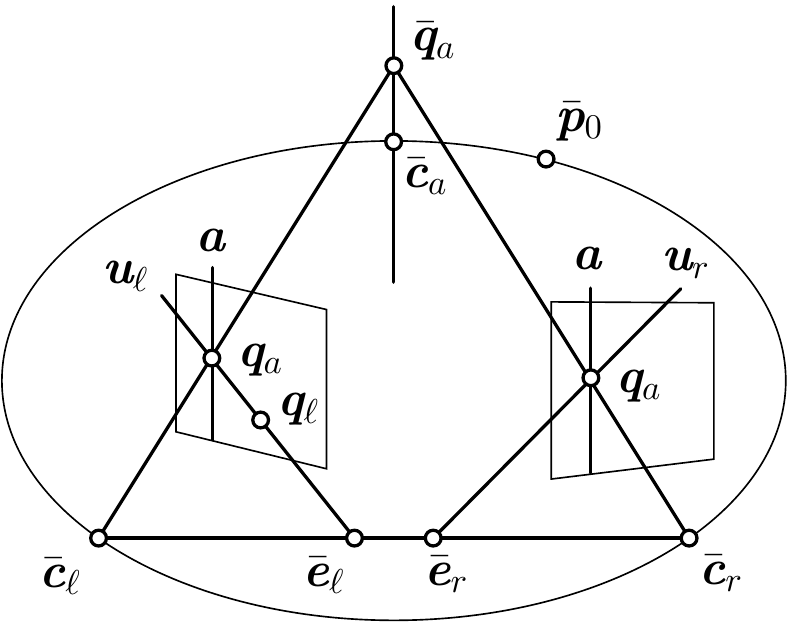}
\caption{Construction of the epipolar geometry. Point $\ivec{q}_\lf$ is given, so the epipolar line 
in $\mathcal{I}_\lf$ is $\ivec{u}_\lf = \ivec{q}_\lf \times \ivec{e}_\lf$. This line intersects the 
image $\ivec{a}$ of the midline horopter in $\mathcal{I}_\lf$ at $\ivec{q}_a = \ivec{a}\times\ivec{u}_\lf$.
The point $\ivec{q}_a$ is on $\ivec{a}$, and is therefore fixed, having the same coordinates $\ivec{a}$ in 
$\mathcal{I}_\rt$. It follows that the epipolar line in $\mathcal{I}_\rt$ is
$\ivec{u}_\rt = \ivec{e}_\rt \times\ivec{q}_a$. The location of $\ivec{q}_\rt$, which corresponds to
$\ivec{q}_\lf$, is unknown, but it must lie on $\ivec{u}_\rt$. The Vieth-M\"{u}ller circle is shown in
the figure, as is the midline-horopter, which passes through $\bar{\ivec{c}}_a$ and $\bar{\ivec{q}}_a$.}
\label{fig:epipolar}
\end{center}
\end{figure}

Suppose that the point $\ivec{q}_\lf$ is given; then, with reference to fig.\ \ref{fig:epipolar},
$\ivec{u}_\lf = \ivec{e}_\lf \times \ivec{q}_\lf$. Furthermore, this line intersects 
the projection $\ivec{a}$ of the midline horopter (\ref{eqn:a}) at the image point 
$\ivec{q}_a = \ivec{a} \times \ivec{u}_\lf$. Any point on $\ivec{u}_\lf$ must also be on
$\ivec{u}_\rt$, as the two lines are images of the same ray. Hence $\ivec{u}_\rt$ can be constructed from $\ivec{e}_\rt$ and the point
in $\mathcal{I}_\rt$ that corresponds to $\ivec{q}_a$. Furthermore, $\ivec{q}_a$ is a 
fixed point (being on $\ivec{a}$), and so its coordinates are unchanged in $\mathcal{I}_\rt$.
It follows that
$\ivec{u}_\rt = \ivec{e}_\rt \times \ivec{q}_a$. The preceding construction may be summarized as
\begin{equation}
\ivec{u}_\rt \sim \ivec{e}_\rt \times \bigl(\ivec{a} \times (\ivec{e}_\lf \times \ivec{q}_\lf) \bigr).
\label{eqn:ur}
\end{equation}
This equation will now be put into a more useful form.
Suppose that $\ivec{w} = (x,y,z)^\tp$; then the cross product $\ivec{w} \times \ivec{p}$ can be 
expressed as a matrix-vector multiplication, $(\ivec{w}\times) \ivec{p}$, where
\begin{equation}
\bigl(\ivec{w}\times\bigr) =
\begin{pmatrix}
0 & -z & y \\
z & 0 & -x \\
-y & x & 0 
\end{pmatrix}
\label{eqn:wcross}
\end{equation}
is a $3\times 3$ antisymmetric matrix, constructed from the components of $\ivec{w}$. 
Consider the part of equation (\ref{eqn:ur}) that does not depend on the particular 
choice of point $\ivec{q}_\lf$; the equivalence (\ref{eqn:wcross}) can be used
to express this as a transformation
\begin{equation}
\imat{E} \sim \bigl(\ivec{e}_\rt\times\bigr) \bigl(\ivec{a}\times\bigr) \bigl(\ivec{e}_\lf\times\bigr)
\label{eqn:e}
\end{equation}
which is the $3 \times 3$ \emph{Essential matrix}\cite{longuet-higgins-1981}. Given a point $\ivec{q}_\lf$, the corresponding point 
$\ivec{q}_\rt$ must be on a certain epipolar line $\ivec{u}_\rt$, as described above. This constraint is
expressed via the Essential matrix as
\begin{equation}
\ivec{q}_\rt^\tp \imat{E} \ivec{q}_\lf = 0 \label{eqn:epipolar}
\end{equation}
where 
$
\ivec{u}_\rt \sim \imat{E} \ivec{q}_\lf
$.
The analogous constraint, 
$
\ivec{q}_\lf^\tp \imat{E}^\tp \ivec{q}_\rt = 0
$
applies in the opposite direction, the epipolar line being
$
\ivec{u}_\lf \sim \imat{E}^\tp\! \ivec{q}_\rt
$
in this case.
The epipoles, as described above, are each the image of the `other' camera centre. This means that 
$\ivec{e}_\lf \sim \imat{R}_\lf(\ivec{b})$, and $\ivec{e}_\rt \sim\imat{R}_\rt(-\ivec{b})$, where 
$\ivec{b}$ is the vector between the optical centres. Equations (\ref{eqn:b}), (\ref{eqn:cl,cr}) and (\ref{eqn:projection}) 
can be used to show that the epipoles are simply
\begin{equation}
\ivec{e}_\lf \sim (\cos\beta_\lf,\, 0,\, \sin\beta_\lf)^\tp
\quad\text{and}\quad
\ivec{e}_\rt \sim (-\cos\beta_\rt,\, 0,\, -\sin\beta_\rt)^\tp.
\end{equation}
These equations can be combined with the definition of the geometric midline horopter 
(\ref{eqn:a}), to give a parametric structure to the Essential matrix.
The non-zero terms in the matrix product (\ref{eqn:e}) are found to be
$E_{12} = -E_\rt \sin\beta_\rt$,
$E_{21} = E_\lf \sin\beta_\lf$,
$E_{23} = -E_\lf \cos\beta_\lf$ and
$E_{22} = E_\rt \cos\beta_\rt$, 
where 
$
E_\lf = \cos\beta_\rt\cos\epsilon + \sin\beta_\rt\sin\epsilon
$
and
$
E_\rt = \cos\beta_\lf\cos\epsilon + \sin\beta_\lf\sin\epsilon
$. The factors $E_\lf$ and $E_\rt$ are seen to be the angle-difference expansions of $\cos(\beta_\lf-\epsilon)$
and $\cos(\beta_\rt-\epsilon)$, respectively. Furthermore, by reference to (\ref{eqn:delta},\ref{eqn:epsilon}) the arguments
$\beta_\lf -\epsilon$ and $\beta_\rt-\epsilon$ are equal to $\pm\delta/2$, and so it follows
from the even-symmetry of the cosine function that $E_\lf=E_\rt=\cos(\delta/2)$.
The Essential matrix is defined here as as a homogeneous transformation
(cf.\ \ref{eqn:epipolar}), and so this common scale-factor can be disregarded, which 
leaves
\begin{equation}
\imat{E}
\sim
\begin{pmatrix}
0 & -\sin\beta_\rt & 0 \\
\sin\beta_\lf & 0 & -\cos\beta_\lf\\
0 & \cos\beta_\rt & 0
\end{pmatrix}.
\end{equation}
It is straightforward to verify that $\ivec{E}$ is indeed an Essential matrix, having one 
singular-value equal to zero, and two identical non-zero singular-values (here equal to unity). 

The same result can be obtained from a variant of the more traditional definition of the essential 
matrix,\cite{longuet-higgins-1981,brooks-1998}
$
\imat{E} \sim \imat{R}_\rt (\ivec{b}\times) \imat{R}_\lf^\tp
$. This definition is, however, not specific to the case of visual fixation, and offers correspondingly less 
insight into the present configuration.

\section{Binocular Disparity}
\label{sec:binocular-disparity}

It was established in the preceding section that projected points in the left and right images must obey the epipolar constraint. In particular it was shown that, given a point $\ivec{q}_\lf$ in $\mathcal{I}_\lf$, the corresponding point $\ivec{q}_\rt$ must lie on a line $\ivec{u}_\rt \sim \imat{E}\ivec{q}_\lf$ in the other image, $\mathcal{I}_\rt$. The structure of the scene displaces the left and right image points along the corresponding lines, which pass through the left and right epipoles, respectively. These image displacements are quantified in this section. 
In particular, it is shown that a scene-point $\bar{\ivec{q}}$ which is in Cyclopean direction $\ivec{p}_c$ will be projected to corresponding image-points $\ivec{q}_\lf$ and $\ivec{q}_\rt$, where
\begin{equation}
\ivec{q}_\lf = \ivec{p}_\lf + t_\lf(s) \, \ivec{d}_\lf
\quad\text{and}\quad
\ivec{q}_\rt = \ivec{p}_\rt + t_\rt(s) \, \ivec{d}_\rt.
\label{eqn:cyc-disparity}
\end{equation}
The unit-vectors $\ivec{d}_\lf$ and $\ivec{d}_\rt$ point towards the respective epipoles $\ivec{e}_\lf$ and 
$\ivec{e}_\rt$. The relation to the epipolar geometry developed in section \ref{sec:epipolar-geometry} is that 
$\ivec{q}_\lf$ and $\ivec{p}_\lf$ are on the same epipolar line; hence, in addition to 
$\ivec{q}_\rt^\tp\imat{E}\ivec{q}_\lf = 0$, as in
(\ref{eqn:epipolar}),
it is true that $\ivec{q}_\rt^\tp\imat{E}\ivec{p}_\lf = 0$.
The formulation (\ref{eqn:cyc-disparity}) makes an approximate correspondence between points $\ivec{p}_\lf$ and 
$\ivec{p}_\rt$, which is corrected by parallax functions $t_\lf(s)$ and $t_\rt(s)$. The common parameter
$s$ is the signed orthogonal distance of $\bar{\ivec{q}}$ from the plane $\mathcal{P}$ which passes through 
the fixation point $\bar{\ivec{p}}_0$. This representation makes it possible, in principle, to estimate a
`local' Cyclopean depth map $S(\ivec{p}_c; \beta, \rho)$ at each fixation point, where $\ivec{p}_c$
ranges over the Cyclopean field of view.

The decomposition (\ref{eqn:cyc-disparity}) has three important properties. Firstly, the unknown parallax 
variables are scalars; there is no need to consider horizontal and vertical disparities separately. Secondly, 
each image correspondence is parameterized by a single variable $s$, which has a direct interpretation as a 
Euclidean distance in the scene. Thirdly, for points close to the fixation plane, the predictions $\ivec{p}_\lf$ 
and $\ivec{p}_\rt$ will be close to $\ivec{q}_\lf$ and $\ivec{q}_\rt$ respectively. In particular, the predicted
correspondence will be exact if the point $\bar{\ivec{q}}$ lies in the fixation plane; $t_\lf(0) = t_\rt(0) = 0$. 
The decomposition (\ref{eqn:cyc-disparity}) will now be described in detail.

The fixation plane $\mathcal{P}$, by definition, passes through the fixation point $\bar{\ivec{p}}_0$, and is 
perpendicular to the Cyclopean gaze direction. Hence the plane has an outward normal vector $\ivec{v}$, as
defined in (\ref{eqn:v}). The plane consists of scene points in the set
\begin{equation}
\mathcal{P} = \Bigl\{\bar{\ivec{p}} : \ivec{v}^\tp(\bar{\ivec{p}}-\bar{\ivec{p}}_0) = 0 \Bigr\}.
\end{equation}
The orthogonal distance from $\mathcal{P}$ to the Cyclopean origin is equal to the range, $\rho$,
of the fixation point, as defined in (\ref{eqn:rho}). The orthogonal distance from $\mathcal{P}$ to 
the scene point $\bar{\ivec{q}}$ will be $s$; hence
\begin{align}
\rho &= \ivec{v}^\tp\bar{\ivec{p}}_0 \label{eqn:rho-vp} \\
s &= \ivec{v}^\tp(\bar{\ivec{q}} - \bar{\ivec{p}}_0). \label{eqn:s}
\end{align}
The range, $\rho$, is strictly positive, whereas $s$ is negative, positive or zero, according
to whether $\bar{\ivec{q}}$ is closer than, further than, or in the plane $\mathcal{P}$, respectively.
Note that $s$ represents the \emph{structure} of the scene with respect to $\mathcal{P}$.
Equations (\ref{eqn:rho-vp}) and (\ref{eqn:s}) can now be used to decompose the Cyclopean depth $z_c$ of 
the point $\bar{\ivec{q}}$ as follows;
\begin{equation}
z_c = \ivec{v}^\tp \bar{\ivec{q}} = \rho + s.
\label{eqn:zc}
\end{equation}
The Cyclopean ray through $\bar{\ivec{q}}$ intersects the fixation plane at $\bar{\ivec{p}}$.
Hence the Cyclopean coordinates of $\bar{\ivec{q}}$ can be expressed as $z_c\ivec{p}_c = \imat{R}\bar{\ivec{q}}$, 
where $\ivec{p}_c$ has been normalized such that $(\ivec{p}_c)_3 = 1$, and $\imat{R}$ encodes the orientation
of the Cyclopean eye (defined by the angle $\beta$, cf.\ equation \ref{eqn:r-l}). The scene-coordinates of points on the corresponding visual ray, parameterized by $z_c$,
can be obtained by inverting this equation. In particular, the intersection $\bar{\ivec{p}}$ of the ray with 
the fixation plane $\mathcal{P}$ can be obtained, as can the original scene-point $\bar{\ivec{q}}$;
\begin{align}
\bar{\ivec{p}} &= \rho \imat{R}^\tp\! \ivec{p}_c\\
\bar{\ivec{q}} &= z_c \imat{R}^\tp\! \ivec{p}_c.
\end{align}
These two points will now be projected into the left image, $\mathcal{I}_\lf$, and the difference
$t_\lf(s)\ivec{d}_\lf$ between the two projections will be evaluated. The analogous derivation applies to the other image, $\mathcal{I}_\rt$, with subscripts `$\lf$' and `$\rt$' exchanged. 
The left coordinates are
\begin{align}
\rho_\lf\ivec{p}_\lf &= \rho\imat{R}_\lf\imat{R}^\tp\! \ivec{p}_c + \tfrac{1}{2}\ivec{e}_\lf \label{eqn:backproj-p}\\
z_\lf \ivec{q}_\lf &= z_c\imat{R}_\lf\imat{R}^\tp\! \ivec{p}_c + \tfrac{1}{2}\ivec{e}_\lf\label{eqn:backproj-q}
\end{align}
where $\tfrac{1}{2}\ivec{e}_\lf = \imat{R}_\lf(-\ivec{c}_\lf)$, as the left, right and Cyclopean optical centres are collinear.
Note that the image-points are normalized such that $(\ivec{p}_\lf)_3 = (\ivec{q}_\lf)_3=1$, 
as in the case of $\ivec{p}_c$. It is necessary to consider the third components of the vectors $\imat{R}_\lf\imat{R}^\tp\!\ivec{p}_c$ and $\tfrac{1}{2}\ivec{e}_\lf$ in (\ref{eqn:backproj-p}) and (\ref{eqn:backproj-q}) respectively; these are
\begin{align}
\lambda_\lf &= (\imat{R}_\lf\imat{R}^\tp \ivec{p}_c)_3 = x_c \sin(\beta_\lf-\beta) + \cos(\beta_\lf-\beta) \label{eqn:lambda} \\
\mu_\lf &=  (\tfrac{1}{2}\ivec{e}_\lf)_3 = \tfrac{1}{2} \sin\beta_\lf \label{eqn:mu}.
\end{align}
The actual image point $\ivec{q}_\lf$ will now be decomposed into a sum of the predicted point $\ivec{p}_\lf$, plus a scalar \emph{parallax} in the direction of a unit vector $\ivec{d}_\lf$. The vector $\ivec{d}_\lf$ should be in the direction of $\ivec{q}_\lf$ with respect to the epipole $\ivec{e}_\lf$. Furthermore, the vector $\ivec{d}_\lf$ must lie in the image plane, $(\ivec{d}_\lf)_3 = 0$. However, it is desirable to avoid defining $\ivec{d}_\lf$ from $\ivec{p}_\lf - \tfrac{1}{2}\ivec{e}_\lf/\mu_\lf$, because $\mu_\lf=0$ whenever $\beta_\lf=0$, as is clear from equation (\ref{eqn:mu}). Hence it is better to use
\begin{equation}
\ivec{d}_\lf = \bigl(\mu_\lf \ivec{p}_\lf - \tfrac{1}{2}\ivec{e}_\lf\bigr) / \kappa_\lf
\label{eqn:d}
\end{equation}
where $(\mu_\lf \ivec{p}_\lf)_3 = (\tfrac{1}{2}\ivec{e}_\lf)_3 = \mu_\lf$, and hence 
$(\ivec{d}_\lf)_3 = 0$. The scalar 
$\kappa_\lf = \bigl|\mu_\lf \ivec{p}_\lf - \tfrac{1}{2}\ivec{e}_\lf\bigr|$ has been introduced, 
so that $\ivec{d}_\lf$ is a unit vector. This is not strictly necessary, but has the advantage  of imposing the original unit of measurement, $|\ivec{b}|$, on the parallax function $t_\lf(s)$ that is associated with each scene point.
The function $t_\lf(s)$ will now be derived. It follows from (\ref{eqn:backproj-p}) and (\ref{eqn:backproj-q}), along with the requirement $(\ivec{p}_\lf)_3 = (\ivec{q}_\lf)_3=1$, that the depth variables $\rho_\lf$ and $z_\lf$ can be expressed as affine functions of the corresponding Cyclopean parameters $\rho$ and $z_c$;
\begin{align}
\rho_\lf &= \lambda_\lf \rho + \mu_\lf \label{eqn:rhol} \\
z_\lf &= \lambda_\lf z_c + \mu_\lf \label{eqn:zl}
\end{align}
where $\lambda_\lf$ and $\mu_\lf$ are the scalars identified by (\ref{eqn:lambda}) and (\ref{eqn:mu}).
A solution for $\lambda_\lf$ can be obtained from either of these equations, and
substituted into the other. The resulting expression can then be solved for $\mu_\lf$;
\begin{equation}
\mu_\lf = \frac{\rho_\lf z_c - \rho z_\lf}{z_c - \rho}.
\label{eqn:mu-depth}
\end{equation}
The equation (\ref{eqn:mu-depth}) is independent of the point $\bar{\ivec{q}}$ that is associated with depths
$z_c$ and $z_\lf$.
The result $\ivec{q}_\lf = \ivec{p}_\lf + t_\lf(s) \, \ivec{d}_\lf$, as in (\ref{eqn:cyc-disparity}), can 
now be derived in full.
Equations (\ref{eqn:backproj-p}) and (\ref{eqn:backproj-q}) are used to express the actual projection 
$\ivec{q}_\lf$ as a function of the predicted projection $\ivec{p}_\lf$;
\begin{equation}
\rho z_\lf \ivec{q}_\lf = \rho_\lf z_c \ivec{p}_\lf - (z_c - \rho) \tfrac{1}{2} \ivec{e}_\lf.
\label{eqn:qp-relation}
\end{equation}
The quantity $\rho z_\lf \ivec{p}_\lf$ is now subtracted from both sides of (\ref{eqn:qp-relation}), and the resulting equation is re-arranged as follows;
\begin{align}
\rho z_\lf (\ivec{q}_\lf - \ivec{p}_\lf) &= 
(\rho_\lf z_c - \rho z_\lf) \ivec{p}_\lf - (z_c - \rho) \tfrac{1}{2} \ivec{e}_\lf \nonumber \\[.75ex]
&= (z_c - \rho) \left(\frac{\rho_\lf z_c - \rho z_\lf}{z_c - \rho}\, \ivec{p}_\lf - \tfrac{1}{2}\ivec{e}_\lf \right) \nonumber \\[.75ex]
&= (z_c - \rho) \bigl(\mu_\lf \ivec{p}_\lf - \tfrac{1}{2} \ivec{e}_\lf\bigr), 
\label{eqn:parallax-1}
\end{align}
where the substitution of $\mu_\lf$ has been made with reference to equation (\ref{eqn:mu-depth}).
Both sides of (\ref{eqn:parallax-1}) are now divided by $\rho z_\lf$, and comparison with (\ref{eqn:d}) leads to
\begin{align}
\ivec{q}_\lf - \ivec{p}_\lf &= \frac{z_c - \rho}{\rho z_\lf}\, \bigl(\mu_\lf \ivec{p}_\lf - \tfrac{1}{2} \ivec{e}_\lf\bigr) \nonumber \\
&= \frac{\kappa_\lf(z_c - \rho)}{\rho z_\lf} \, \ivec{d}_\lf \nonumber \\
&= \frac{\kappa_\lf s}{\rho z_\lf} \, \ivec{d}_\lf
\label{eqn:parallax-2}
\end{align}
where (\ref{eqn:zc}) has been used above, to make the substitution 
$s = z_c - \rho$. 
The practical problem with (\ref{eqn:parallax-2}) is that in addition to the free parameter $s$, 
the variable $z_\lf$ is apparently unknown. This is resolved by making the substitution $z_\lf = \lambda_\lf (\rho + s) + \mu_\lf$, 
which follows from (\ref{eqn:zc}) and (\ref{eqn:zl}).
Therefore, if $\ivec{p}_\lf$ is added to both sides of 
(\ref{eqn:parallax-2}), then the result is 
\begin{gather}
\ivec{q}_\lf = \ivec{p}_\lf + t_\lf(s) \, \ivec{d}_\lf \label{eqn:parallax-3a} \\[1ex]
\text{with} \nonumber \\
t_\lf(s) = \frac{\kappa_\lf (s/\rho)}{\lambda_\lf(\rho+s)+\mu_\lf}.
\label{eqn:parallax-3b}
\end{gather}
From this it can be seen that the function $s \mapsto t_\lf$ given by (\ref{eqn:parallax-3b}) is a
\oned projective transformation, from which it may be inferred that $t_\lf$ and $t_\rt$
are also related by a projective transformation. 
The analogous definitions are made for $\mathcal{I}_\rt$, with subscripts `$\lf$' and `$\rt$' exchanged.
Equations (\ref{eqn:parallax-2}), (\ref{eqn:parallax-3a}) and (\ref{eqn:parallax-3b}) can be interpreted as follows. It is assumed that the Cyclopean coordinates $(\ivec{v},\rho)$ of the fixation point $\bar{\ivec{p}}_0$ are known. Then, given
the Cyclopean direction $\ivec{p}_c$ of another point, it is possible to compute the predicted point $\ivec{p}_\lf$ (\ref{eqn:backproj-p}), as well as the vector $\kappa_\lf\ivec{d}_\lf$ (\ref{eqn:d}). The scalars $\lambda_\lf$ and $\mu_\lf$ are obtained from (\ref{eqn:lambda}) and (\ref{eqn:mu}), respectively.
The unknown parallax, $t_\lf(s)$ is proportional to $s/z_\lf$; this ratio is the depth of the scene point $\bar{\ivec{q}}_\lf$ with respect to the fixation plane $\mathcal{P}$, divided by the depth of the point with respect to the left viewing direction. 

For points that are on the fixation plane, $s=0$, and therefore it is clear from (\ref{eqn:parallax-3b}) 
that $t_\lf(s)=0$. It follows from (\ref{eqn:cyc-disparity}) that $\ivec{q}_\lf = \ivec{p}_\lf$ and 
$\ivec{q}_\rt = \ivec{p}_\rt$. This makes it interesting to consider the relationship between 
$\ivec{p}_\lf$ and $\ivec{p}_\rt$. It can be shown that the points $\ivec{p}_\lf$ can be mapped onto 
the corresponding points $\ivec{p}_\rt$ by a projective transformation
\begin{equation}
\ivec{p}_\rt = \imat{H} \ivec{p}_\lf.
\label{eqn:hp}
\end{equation}
The $3\times 3$ matrix $\imat{H}$ is the \emph{homography} induced by the fixation plane $\mathcal{P}$. 
If $w_\lf$ is the perpendicular distance from $\bar{\ivec{c}}_\lf$ to $\mathcal{P}$, then the
transformation has the form
\begin{equation}
\imat{H} = \imat{R}_\rt 
\biggl(\imat{I} - \frac{\ivec{b} \ivec{v}^\tp}{w_\lf}\biggr)
\imat{R}_\lf^\tp.
\label{eqn:h}
\end{equation}
In the general case, $s \ne 0$, equations (\ref{eqn:cyc-disparity}), (\ref{eqn:hp}) and (\ref{eqn:h}) 
can be combined, leading to the well-known `plane plus parallax' decomposition\cite{shashua-navab-1996,luong-1996}
\[
\ivec{q}_\rt = \imat{H} \ivec{p}_\lf + t_\rt(s) \ivec{d}_\rt.
\]
The symmetric representation (\ref{eqn:cyc-disparity}) is, however, preferable in the present
context. This is because it encodes the depth map directly in Cyclopean coordinates, 
$S(\ivec{p}_c; \beta, \rho)$.

\begin{figure}[!ht]
\begin{center}
\includegraphics[scale=1.0]{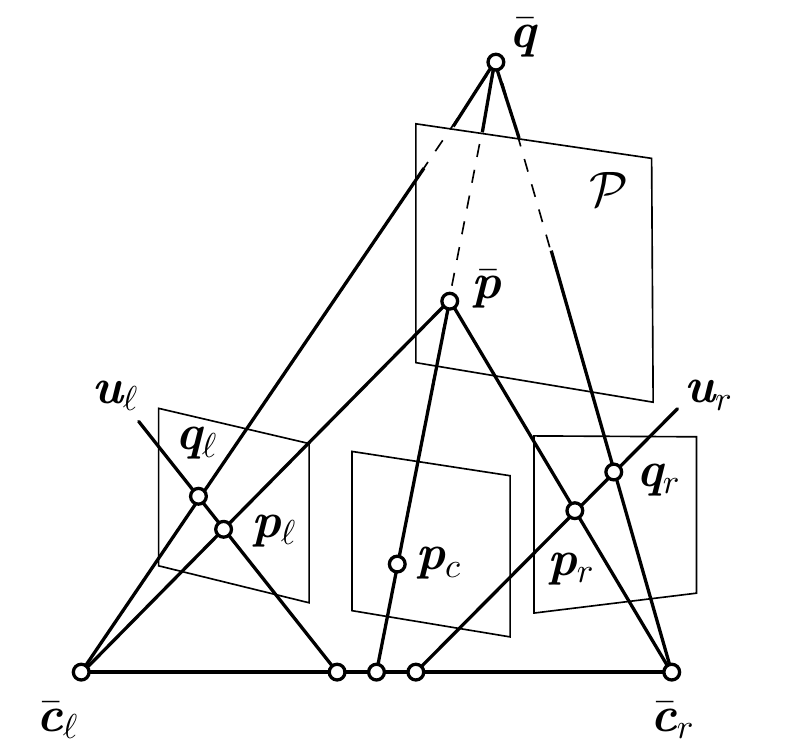}
\caption{Geometry of Cyclopean parallax. The fixation plane $\mathcal{P}$ is defined by the fixation point $\bar{\ivec{p}}_0$, and is parallel to the Cyclopean image plane. The point $\ivec{p}_c$ defines a Cyclopean ray, which 
intersects the fixation plane at $\bar{\ivec{p}}$, and the scene at $\bar{\ivec{q}}$. The scene point has depth $s$ with
respect to $\mathcal{P}$. The predicted image-projections of $\bar{\ivec{q}}$ are at $\ivec{p}_\lf$ and $\ivec{p}_\rt$.
The true projections, $\ivec{q}_\lf$ and $\ivec{q}_\rt$, are displaced along the corresponding epipolar lines,
$\ivec{u}_\lf$ and $\ivec{u}_\rt$, respectively. The displacement can be parameterized by $s$, as described 
in the text.}
\label{fig:disparity}
\end{center}
\end{figure}

\section{Discussion}
\label{sec:discussion}

Three problems of binocular vision were identified in section~\ref{sec:introduction}, 
concerning oculomotor parameterization, disparity processing, and scene representation. 
A unified geometric treatment of these problems has been given in 
sections~\ref{sec:binocular-orientation}--\ref{sec:binocular-disparity}. The main psychophysical 
and physiological findings that relate to each of the proposed geometric solutions will 
now be reviewed.

\subsection{Oculomotor Parameterization}

As described in section \ref{sec:binocular-orientation}, the binocular orientation of the 
eyes can be specified, in the visual plane, by the vergence and version angles, $\delta$ and
$\epsilon$. Furthermore, these variables are convenient for the specification 
of a coordinated eye movements from one fixation point to another. It was suggested
by Hering that the oculomotor system actually encodes binocular eye movements in
terms of vergence and version \cite{hering-1868,carpenter-1988}. Specifically, Hering's law
of \emph{equal innervation} states that the eyes are oriented by
signals $(\delta,\epsilon)$ where, according to (\ref{eqn:delta},\ref{eqn:epsilon}),
$\beta_\lf=\epsilon+\delta/2$ and $\beta_\rt=\epsilon-\delta/2$.
 Each eye moves according to the sum of the 
appropriate vergence and version components (which may cancel; e.g.\ if 
$\epsilon + \delta/2 = 0$, then both the initial and final fixation points
lie along the visual direction $\ivec{v}_\lf$).
The existence of physiological mechanisms that encode pure vergence has 
been demonstrated in the midbrain \cite{mays-1984}, and it has been 
suggested that the vergence/version decomposition is used to represent 
the difference between the current and target fixation points.
However, the actual trajectories of large binocular eye movements are not consistent 
with the simple vergence/version decomposition \cite{collewijn-1997}. Furthermore, 
it has been found that the visual axes can be widely mis-aligned during \textsc{rem}
sleep \cite{zhou-1997}, which seems more consistent with the independent
parameterization of each eye.

It may be that the application of Hering's law is limited to those situations
in which the existence of a \threed fixation point can be ensured by the 
foveal correspondence of the images.
This condition would, for example, distinguish the vergence response
from the disconjugate component of a binocular saccade. This is because 
vergence is driven by visual feedback\cite{rashbass-1961} (alignment of the retinal images),
which is not generally available during the course of a binocular saccade.
Likewise, when the eyes are closed, there is no visual information to ensure 
the existence of a \threed fixation point. In summary, the evidence for Hering's 
law of equal innervation is mixed, but it seems clear that there are situations
in which it is not a satisfactory model.

\subsection{Disparity Processing}

It is possible to separate the initial estimation of image correspondences from 
the \emph{interpretation} of the resulting disparity field; this distinction is
made in several computational models of stereopsis\cite{marr-1976,foley-1980,mayhew-1982}. 
It is supposed, in these models, that the correspondence problem is first solved,
independently of the gaze parameters. The latter are then recovered from the 
estimated disparity field\cite{mayhew-1982}, and the two types of information are
combined, leading to a \threed (though not necessarily Euclidean) interpretation 
of the scene\cite{garding-1995}. This scheme is compatible with the physiological 
basis of stereopsis; for example, it has been demonstrated that the initial 
binocular mechanisms in area~V1 are tuned to absolute disparity\cite{cumming-1999},
as described in section~\ref{sec:introduction-b}. This finding indicates that the 
low-level mechanisms of stereopsis do not `compensate' for any disparity that is 
imposed by the relative orientation of the eyes.

The biological feasibility of a general solution to the binocular correspondence 
problem will now be considered. Individual disparity-tuned cells, in area~V1, 
typically respond to a small range of absolute disparities, centred on some preferred value.
The distribution of preferred disparities, over the population of cells, can
also be inferred from the experimental data\cite{anzai-1999}. This arrangement
suggests the following difficulty: It seems likely that, for any particular scene, 
and any particular fixation point, a large proportion of the V1 disparity response 
may be spurious. This is because the occurence of the preferred absolute disparity,
for a given detector, effectively depends on the orientation of the eyes, as well
as on the scene structure. Hence the possibility of detecting a `false match' is
exacerbated by the fact that, owing to the variable orientation of the eyes, the 
true match may not even be in the range of a given detector.

One way to address this problem is by imposing some prior knowledge about the 
typical structure of the scene. For, example, it might be assumed that the 
scene is approximately planar in the neighbourhood of the fixation point.
Such a model could then be used to define \emph{sets} of disparity detectors
that are effectively tuned to the same surface in depth. This is done by 
computing the image-to-image mapping induced by the surface model, and comparing
it to the disparity response. Note that this process is entirely consistent with
detectors that respond to absolute disparity, as the scene model does not influence 
the output of the individual mechanisms. Rather, the local scene model is 
\emph{imposed} on the collective response.

If this approach is to be effective, then the induced image-to-image mapping 
must be appropriately parameterized. It is important to note that an appropriate 
parameterization would allow prior knowledge to be imposed on the gaze parameters, 
as well as on the scene structure. For example, image-to-image mappings 
associated with extreme configurations of the eyes might be penalized, and
some mappings might be excluded altogether (e.g.\ those associated with
non-intersecting visual axes). It is emphasized that this approach does
not require oculomotor information about the current gaze parameters; rather, 
it assumes that a model of gaze-variation can be learned from the image data.
The geometric constraints described in sections~\ref{sec:binocular-orientation} 
and~\ref{sec:epipolar-geometry} would seem to make this a biologically
feasible task. For example, if the scene is assumed to be approximately 
perpendicular to the Cyclopean gaze direction, then the appropriate
scene/gaze model has just three parameters; $(\alpha,\beta,\rho)$. 
Furthermore, the form of the induced image-to-image mapping has been 
exactly described in section~\ref{sec:binocular-disparity}.

% dobbins 
% trotter

\subsection{Scene Representation}

It was shown in section~\ref{sec:binocular-disparity} that binocular disparity can 
conveniently be measured with respect to the fixation plane, $\mathcal{P}$. This plane passes 
through the fixation point, and is orthogonal to the Cyclopean viewing direction.
This construction is commonly used in physiological studies of stereopsis. In particular, 
the shape of the disparity tuning profile, with respect to the depth of the fixation
plane, has been used to classify binocular neurons\cite{poggio-1977}. Subsequent
experiments have suggested that there exists a continuum of different disparity
tuning curves \cite{prince-2002}, rather than a number of distinct types. 
Nonetheless, the continuum of disparity tuning is clearly organized around the fixation
plane; for example, the majority of cells are tuned to disparities close to that of the 
plane\cite{cumming-2001}.

The importance of the fixation plane is also reflected in psychophysical studies
of stereopsis. As noted in section~\ref{sec:introduction-c}, only those points
in space that are in Panum's area can be binocularly fused \cite{ogle-1950}. It 
has further been demonstrated, using simple stimuli, that stereo acuity is highest
for targets located close to the fixation plane \cite{ogle-1950,blakemore-1970}. 
However, the representation of more complex binocular stimuli raises further
questions. In particular, it has been shown that judgment of the \emph{relative} 
depth between nearby targets is much more accurate than judgment of the deviation
of a singe target from the fixation plane \cite{westheimer-1979}. Furthermore, the 
surface that is perceived in depth is, in some cases, an \emph{interpolation} 
of the point-wise disparity stimulus \cite{mitchison-1985}. These observations
suggest that the representation of binocular disparity may depend on the 
local scene-structure, as well as on the gaze parameters \cite{glennerster-2002}.
The present model is compatible with this approach, and indeed, the fixation 
plane $\mathcal{P}(\rho,\ivec{v})$ can be interpreted as the 
zeroth-order approximation of the local scene structure. It is straightforward to 
substitute the first-order model $\mathcal{P}(\rho,\ivec{v}; \theta,\phi)$, 
in which the angles $\theta$ and $\phi$ represent the local surface orientation, 
and to repeat the derivation of binocular disparity in section~\ref{sec:binocular-disparity}.

%The drawback of this extension is that errors in the estimated parameters 
%$(\theta,\phi)$ would distort the resulting depth-map.

The integration of the visual information across the larger scene will now be
considered. The representation described in section~\ref{sec:binocular-disparity}
allows the estimated viewing distance to be combined directly with the
Cyclopean depth map, because both are measured along the Cyclopean gaze direction,
as in equations (\ref{eqn:rho-vp}) and (\ref{eqn:s}). The global structure of 
the scene could therefore be encoded in a collection of local depth-maps, of the type 
described above. Although each depth-map would be associated with a different fixation
point, it would be geometrically straightforward to combine the encodings, based on
the corresponding gaze parameters.

Finally, it can be argued that the Cyclopean representation is consistent
with the perception of visual space\cite{hering-1868}. Human binocular
vision results, at least subjectively, in a single view of the scene. If this
synthetic view has a meaningful centre of projection, then it may be hypothesized 
that it is located at the Cyclopean point\cite{ono-1995}.

\subsection{Conclusions}

A Cyclopean parameterization of binocular vision has been developed in detail. 
The parameterization has been used to construct the horopter and the epipolar 
geometry of a fixating visual system. Furthermore, the effect of the oculomotor 
parameters on the binocular disparity field has been described. It is clear that
the interpretation of the disparity field is complicated by the variable 
orientation of the eyes. However, it has been argued here that this complication 
is minimized by the binocular coordination of the eyes. The geometric and 
computational appeal of the Cyclopean representation has been emphasized,
and the biological relevance of the model has been indicated.


\begin{thebibliography}{}

\bibitem{anzai-1999}
A.\ Anzai, I.\ Ohzawa \& R.\ D.\ Freeman.
``Neural Mechanisms for Encoding Binocular Disparity: Receptive Field Position versus Phase.'' 
Journal of Neurophysiology 82, 
874--890 (1999).

\bibitem{armstrong-1995}
M.\ Armstrong, A.\ Zisserman \& R.\ I.\ Hartley.
``Self-Calibration from Image Triplets.'' Proc.\ 4th European Conference on Computer Vision, 
vol.\ 1, p.\ 3--16 (1996).

\bibitem{bergen-1992}
J.\ R.\ Bergen, P.\ Anandan, K.\ J.\ Hanna \& R.\ Hingorani.
``Hierarchical Model-Based Motion Estimation''.
Proc.\ 2nd European Conference on Computer Vision, 
237--252 (1992).

\bibitem{blakemore-1970}
C.\ Blakemore.
``The Range and Scope of Binocular Depth Discrimination in Man.''
Journal of Physiology 211, 
599--622 (1970).

\bibitem{brooks-1998}
M.\ J.\ Brooks, L.\ de Agapito, D.\ Q.\ Huynh \& L. Baumela.
``Towards Robust Metric Reconstruction via a Dynamic Uncalibrated Stereo Head.''
Image and Vision Computing, 16, 
989--1002 (1998).

\bibitem{carpenter-1988}
R.\ H.\ S.\ Carpenter.
\emph{Movements of the Eyes.} (Pion 1988).

\bibitem{collewijn-1997}
H.\ Collewijn, C.\ J.\ Erkelens \& R.\ M.\ Steinman.
``Trajectories of the Human Binocular Fixation Point during Conjugate and Non-Conjugate Gaze-Shifts.''
Vision Research 37(8), 
1049--1069 (1997).

\bibitem{cooper-1979}
M.\ L.\ Cooper \& J.\ D.\ Pettigrew.
``A Neurophysiological Determination of the Vertical Horopter in the Cat and Owl.''
J.\ Comparative Neurology 184,
1--25 (1979).

\bibitem{cumming-1999}
B.\ G.\ Cumming \& A.\ J.\ Parker. 
``Binocular Neurons in V1 of Awake Monkeys are Selective for Absolute, not Relative Disparity.''
Journal of Neuroscience 19, 
5602--5618 (1999).

\bibitem{cumming-2001}
B.\ G.\ Cumming \& G.\ C.\ DeAngelis.
``The Physiology of Stereopsis.''
Annual Review of Neuroscience 24,
203--238 (2001).

\bibitem{faugeras-1995} O.\ Faugeras. 
``Stratification of \threed Vision: Projective, Affine, and Metric Representations.'' 
\josaa 12(3), 
465--484 (1995).

\bibitem{fleet-1996}
D.\ J.\ Fleet, H.\ Wagner \& D.\ J.\ Heeger.
``Neural Encoding of Binocular Disparity: Energy Models, Position Shifts and Phase Shifts.''
Vision Research 36(12), 
1839--1857 (1996).

\bibitem{foley-1980}
J.\ M.\ Foley.
``Binocular Distance Perception''. 
Psychological Review 87, 
411-434 (1980).

\bibitem{garding-1995}
J.\ Garding, J.\ Porrill, J.\ E.\ W.\ Mayhew, \& J.\ P.\ Frisby.
``Stereopsis, Vertical Disparity and Relief Transformations''.
Vision Research 35(5), 
703--722 (1995).

\bibitem{glennerster-2002}
A.\ Glennerster, S.\ P.\ McKee, \& M.\ D.\ Birch. 
``Evidence for Surface-Based Processing of Binocular Disparity''. 
Current Biology 12,
825--828 (2002).

\bibitem{helmholtz-1910}
H.\ L.\ F.\ von Helmholtz. 
\emph{Treatise on Physiological Optics, Vol.\ III} 
(Optical Society of America, 3rd ed.\ 1910, trans.\ J.\ P.\ C.\ Southall, 1925).

\bibitem{hepp-1994}
Hepp, K. (1994).
``Oculomotor Control: Listing's Law and All That.'' 
Current Opinion in Neurobiology 4,
862--868.

\bibitem{hering-1868}
E.\ Hering.
\emph{The Theory of Binocular Vision}
(Englemann 1868; ed.\ \& trans.\ B.\ Bridgeman \& L.\ Stark, Plenum Press, 1977).

\bibitem{julesz-1971}
B.\ Julesz. \emph{Foundations of Cyclopean Perception.} 
(University of Chicago Press, 1971).

\bibitem{julesz-1972}
B.\ Julesz.
``Cyclopean Perception and Neurophysiology''.
Investigative Ophthalmology and Visual Science 11, 
540--548 (1972).

\bibitem{koenderink-1976}
J.\ J.\ Koenderink \& A.\ J.\ van Doorn. 
``Geometry of Binocular Vision and a Model for Stereopsis''
Biological Cybernetics 21(1), 
29--35 (1976).

\bibitem{longuet-higgins-1981}
H.\ C.\ Longuet-Higgins. 
``A Computer Algorithm for Reconstructing a Scene from Two Projections.'' 
Nature 293, 
133--135 (1981).

\bibitem{luong-1996}
Q.-T.\ Luong \& T.\ Vi\'{e}ville.
``Canonical Representations for the Geometries of Multiple Projective Views.''
Computer Vision and Image Understanding 64(2), 
193--229 (1996).

\bibitem{marr-1976}
D.\ Marr \& T.\ Poggio.
``Cooperative Computation of Stereo Disparity.''
Science 194 (4262), 
283--287 (1976)

\bibitem{mayhew-1982}
J.\ E.\ W.\ Mayhew \& H.\ C.\ Longuet-Higgins.
``A Computational Model of Binocular Depth Perception.''
Nature 297, 
376--379 (1982).

\bibitem{mays-1984}
L.\ E.\ Mays. 
``Neural Control of Vergence Eye Movements: Convergence and Divergence Neurons in Midbrain.''
Journal of Neurophysiology 51 (5), 
1091--1108 (1984).

\bibitem{mitchison-1985}
G.\ J.\ Mitchison \& S.\ P.\ McKee.
``Interpolation in Stereoscopic Matching.''
Nature 315, 
402--404 (1985).

\bibitem{ogle-1950}
K.\ N.\ Ogle.
\emph{Researches in Binocular Vision} 
(W.\ B.\ Saunders, 1950).

\bibitem{ono-1995}
H.\ Ono \& A.\ P.\ Mapp.
``A Re-Statement and Modification of Wells-Hering's Laws of Visual Direction.''
Perception 24(2), 
237--252 (1995).

\bibitem{pettigrew-1979}
J.\ D.\ Pettigrew, 
``Binocular Visual Processing in the Owl's Telencephalon.''
Proc.\ Royal Society of London B-204, 
435--454 (1979).

\bibitem{pettigrew-1986} 
J.\ D.\ Pettigrew, ``Evolution of Binocular Vision.''  
In \emph{Visual Neuroscience}, eds.\ J.\ D.\ Pettigrew, K.\ J.\ Sanderson \& W.\ R.\ Levick.
208--22 (Cambridge University Press. 1986).

\bibitem{poggio-1977}
G.\ F.\ Poggio \& B.\ Fischer.
``Binocular Interaction and Depth Sensitivity in Striate and Prestriate Cortex of Behaving Rhesus Monkeys.''
Journal of Neurophysiology 40, 
1392--1407 (1977).

\bibitem{prince-2002}
S.\ J.\ D.\ Prince, B.\ G.\ Cumming \& A.\ J.\ Parker.
``Range and Mechanism of Encoding of Horizontal Disparity in Macaque V1.''
Journal of Neurophysiology 87, 
209--221 (2002).

\bibitem{rashbass-1961}
C. Rashbass \& G. Westheimer.
``Disjunctive Eye Movements.''
Journal of Physiology 159, 
339--360 (1961).

\bibitem{shashua-navab-1996}
A.\ Shashua \& N.\ Navab. 
``Relative Affine Structure: Canonical Model for \threed from \twod Geometry and Applications.'' 
IEEE Trans.\ Pattern Analysis and Machine Intelligence 18(9), 
873--883 (1996).

\bibitem{todd-2001}
J.\ T.\ Todd, A.\ H.\ J.\ Oomes, J.\ J.\ Koenderink \& A.\ M.\ L.\ Kappers.
``On the Affine Structure of Perceptual Space.''
Psychological Science 12(3), 
191--196 (2001).

\bibitem{tyler-1991}
C.\ W.\ Tyler.
``The Horopter and Binocular Fusion'' In \emph{Vision and Visual Disorders, vol. 9; Binocular Vision}. 
ed.\ D.\ Regan, 19--37 (Macmillan: New York, pp. 38-74, 1991).

\bibitem{walls-1961}
G.\ L.\ Walls.
``The Evolutionary History of Eye Movements.''
Vision Research 2, 
69--80 (1962).

\bibitem{westheimer-1979}
G.\ Westheimer.
``Cooperative Neural Processes involved in Stereoscopic Acuity.''
Experimental Brain Research 36(3), 
585--597 (1979).

\bibitem{zhou-1997}
W.\ Zhu \& W.\ M.\ King. 
``Binocular Eye Movements Not Coordinated During \textsc{rem} Sleep.''
Experimental Brain Research 117, 
153--160 (1997).

\end{thebibliography}
\end{document}